%% file: main.tex
\documentclass[runningheads]{llncs}

\usepackage{graphicx}
\usepackage{comment}
\usepackage{amsmath,amssymb}
\usepackage{color}
\usepackage{url}
\usepackage[hidelinks]{hyperref}

\usepackage{times}
\usepackage{epsfig}
\usepackage{dirtytalk}

\usepackage[dvipsnames]{xcolor}
\usepackage[symbol]{footmisc}
\usepackage{booktabs}
\usepackage{multirow}

\begin{document}

\title{Distractor-Aware Video Object Segmentation}

\include{00-setup}


\author{Andreas Robinson
\and Abdelrahman Eldesokey
\and Michael Felsberg
}

\authorrunning{A. Robinson et al.}

\institute{Linköping University, Sweden,
\email{\{firstname.lastname\}@liu.se}}

\maketitle              

\renewcommand{\thefootnote}{}
\footnotetext{This version of the contribution has been accepted for publication,
after peer review, but is not the Version of Record and does not reflect
post-acceptance improvements, or any corrections. The Version of Record is
available online at: \url{https://doi.org/10.1007/978-3-030-92659-5_14}.
Use of this Accepted Version is subject to the publisher's Accepted Manuscript
terms of use \url{https://www.springernature.com/gp/open-research/policies/accepted-manuscript-terms}.}
\renewcommand{\thefootnote}{\fnsymbol{footnote}}

\begin{abstract}

Semi-supervised video object segmentation is a challenging task that aims to segment a target throughout a video sequence given an initial mask at the first frame.
Discriminative approaches have demonstrated competitive performance on this task at a sensible complexity.
These approaches typically formulate the problem as a one-versus-one classification between the target and the background.
However, in reality, a video sequence usually encompasses a target, background, and possibly other distracting objects.
Those objects increase the risk of introducing false positives, especially if they share visual similarities with the target.
Therefore, it is more effective to separate distractors from the background, and handle them independently.

We propose a one-versus-many scheme to address this situation by separating distractors into their own class.
This separation allows imposing special attention to challenging regions that are most likely to degrade the performance.
We demonstrate the prominence of this formulation by modifying the learning-what-to-learn \cite{vos_lwl} method to be distractor-aware.
Our proposed approach sets a new state-of-the-art on the \textit{DAVIS 2017 val} dataset, and improves over the baseline on the \textit{DAVIS 2017 test-dev} benchmark by 4.6 percentage points.
\end{abstract}

\section{Introduction}

Semi-supervised video object segmentation (VOS) aims to segment a target throughout a video sequence, given an initial segmentation mask in the first frame. 
This task can be very challenging due to camera motion, occlusion, and background clutter.
Several deep learning based methods have been proposed recently to address these challenges \cite{vos_lwl,vos_rgmp,vos_stm,vos_cvpr,vos_kmn,vos_cfbi}.
Among those, discriminative methods \cite{vos_lwl,vos_cvpr} have shown competitive performance at a reasonable computational cost, making them suitable for real-time applications, 
e.g.,
enhancing visual object tracking in crowded scenes, removing or replacing the background in video sequences or live conference calls, for privacy masking in surveillance videos, or as an attention mechanism in downstream vision tasks such as action recognition.

\setlength{\tabcolsep}{1pt}
\begin{figure}[t!]
\begin{tabular}{ccc}
     LWL \cite{vos_lwl} & \textbf{Ours} & Ground truth \\
     \includegraphics[width=0.31\columnwidth, trim={9cm 1cm 11cm 3cm},clip]{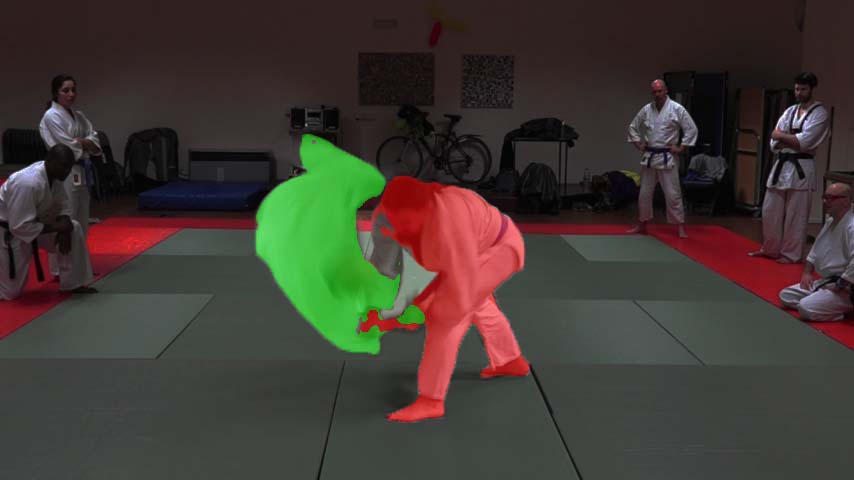} &
     \includegraphics[width=0.31\columnwidth, trim={9cm 1cm 11cm 3cm},clip]{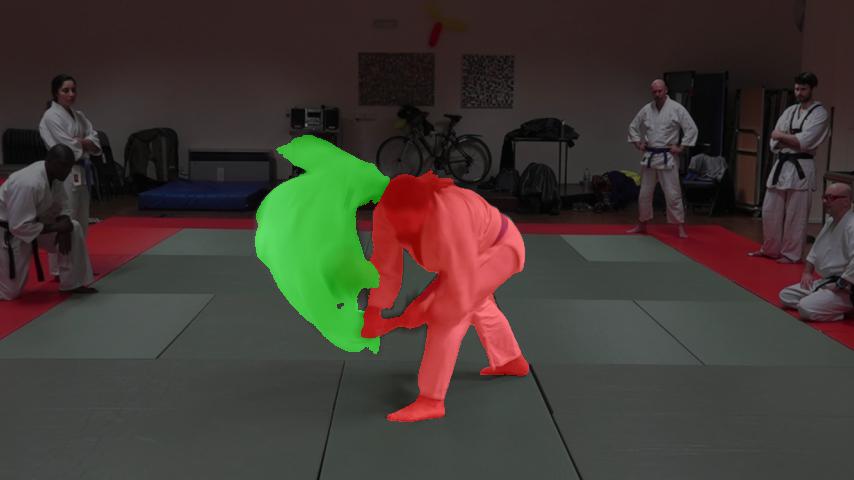} &
     \includegraphics[width=0.31\columnwidth, trim={9cm 1cm 11cm 3cm},clip]{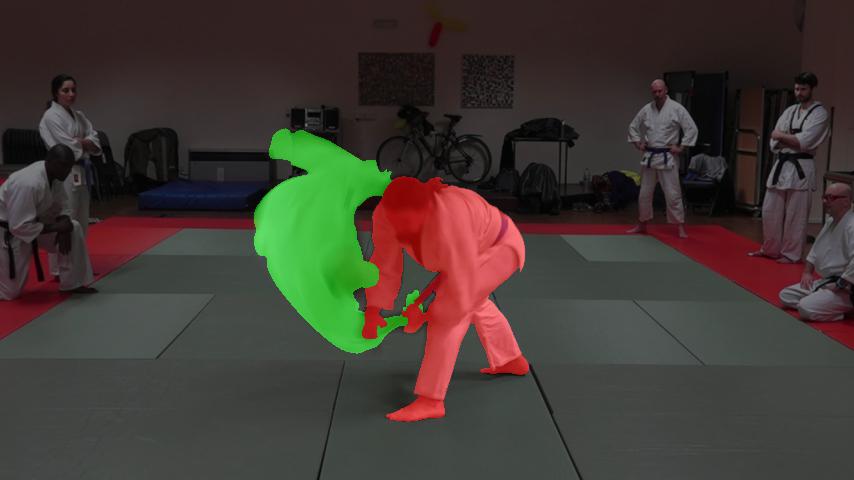}\\
\end{tabular}
\caption{The impact of incorporating distractor-awareness into the baseline (LWL \cite{vos_lwl}).
Our distractor-aware approach produces more accurate predictions than the baseline in highly ambiguous regions, where objects share visual similarities.}
\label{fig:intro}
\end{figure}
\setlength{\tabcolsep}{4pt}

The majority of the discriminative approaches formulate the problem as a one-versus-one classification between the target and the background.
Based on this, they attempt to construct a robust representation of the target that is as distinct as possible from the background.
However, the background usually includes other objects that could be visually similar to the target.
In this case, it might be challenging to find a good representation that discriminates the target from those distracting objects.
This can usually lead to several false positive as the classifier is likely to fail given the underlying coarse representation between the target and the background.
Figure \ref{fig:intro} shows an example where a top-performing discriminative approach fails to discriminate between the target and other objects of the same type.

In this paper, we address this aforementioned challenge by reformulating the problem as a one-versus-many classification.
We propose to separate the distracting objects from the background, and handle them as a distinct class.
As a result, making the network aware of these distractors during training, promotes the learning of a robust representation of the target that is more discriminative against both the background \emph{and} distractors.

We demonstrate the effectiveness of our approach by modifying the learning-what-to-learn (LWL) approach \cite{vos_lwl} to become distractor-aware.
First, we modify the learning pipeline to incorporate information about the distractors.
In case of videos with multiple objects, we initialize other objects in the scene as distractors.
Second, to enhance the discriminative power of the network, we integrate high-resolution features to the target model.
Finally, we introduce the use of adaptive refinement and upsampling \cite{raft} as a regularization to enforce local consistency at uncertain regions such as edges, and  object-to-object boundaries.

Experiments show that our proposed framework sets a new state-of-the-art result on the DAVIS 2017 \textit{val} dataset \cite{dv2017}. Moreover, we improve the results over the baseline on the DAVIS 2017 \textit{test-dev} benchmark by a large margin.
On the YouTube-VOS 2018 dataset \cite{yt2018}, which is characterized by limited annotation accuracy at object-to-distractor boundaries, our method still shows significant improvement over the baseline.

 
The remainder of the paper is organized as follows: we start by providing an overview of existing discriminative VOS approaches in the literature, distractor-awareness in other vision tasks, and existing upsampling and refinement methods in VOS.
Next, we briefly describe the baseline approach, Learning-what-to-learn \cite{vos_lwl}, followed by our proposed distractors modelling.
Finally, we provide quantitative and qualitative results for our proposed approach in comparison with the baseline, and existing state-of-the-art methods as well as an ablation study.




\input{20-related_work}


\input{30-method}


\input{40-experiments}


\section{Conclusion}
We have proposed a distractor-aware, discriminative video object segmentation approach.
In contrast to existing methods, our proposed method encodes distractors into a separate class, to exploit information about other objects in the scene that are likely to be confused with the target.
Moreover, we propose the use of joint refinement and upsampling to regularize the likelihoods for highly uncertain regions with their neighborhoods.
We demonstrated the effectiveness of our approach by modifying an existing state-of-the-art approach to be distractor-aware.
Our modification sets a new state-of-the-art on the DAVIS 2017 \textit{val} dataset, while improving over the baseline with a remarkable margin on the DAVIS 2017 \textit{test-dev} dataset.
These results clearly indicate the efficacy of explicitly modelling distractors when solving video object segmentation.

\section{Acknowledgements}

This project was partially supported by the Wallenberg AI, Autonomous Systems and Software Program (WASP) funded by the Knut and Alice Wallenberg Foundation, the Excellence Center at Linköping-Lund in Information Technology (ELLIIT), the Swedish Research Council grant no.~2018-04673, and the Swedish Foundation for Strategic Research (SSF)  project Symbicloud. 

\newpage


\bibliographystyle{splncs04}
\bibliography{egbib}


\newpage
\appendix

\section{Qualitative Results}
\begin{figure*}[t]
\centering
\tabcolsep=0.01cm
\renewcommand{\arraystretch}{0.06}
\begin{tabular}{ccc}
LWL & \textbf{Ours} & Ground truth \\
&& \\
\includegraphics[width=0.32\textwidth]{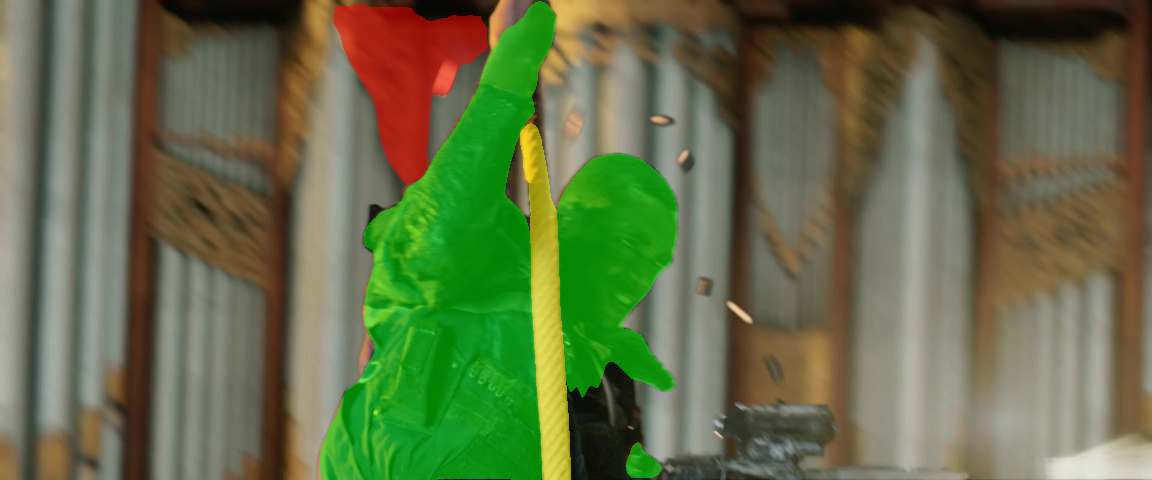} &
\includegraphics[width=0.32\textwidth]{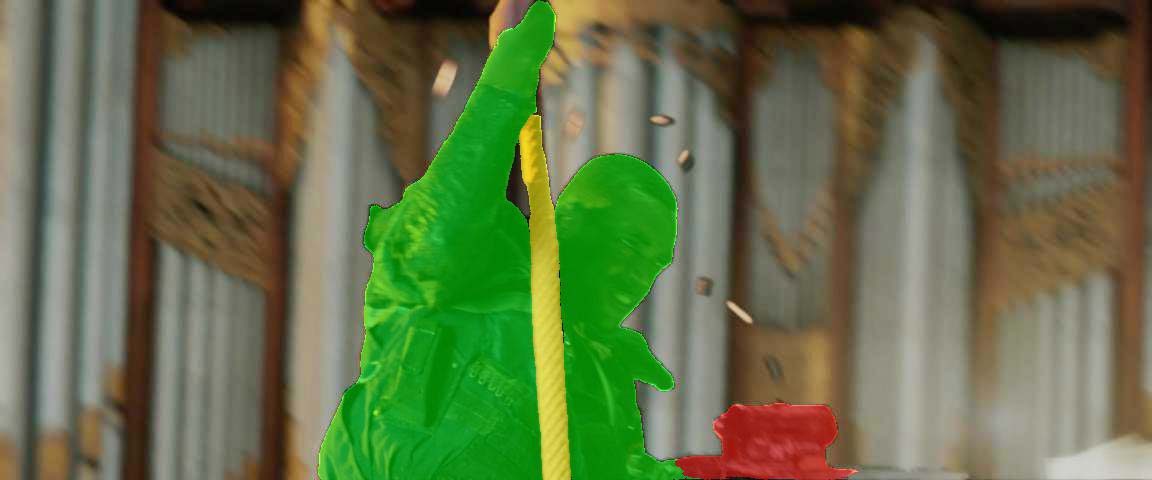} &
\includegraphics[width=0.32\textwidth]{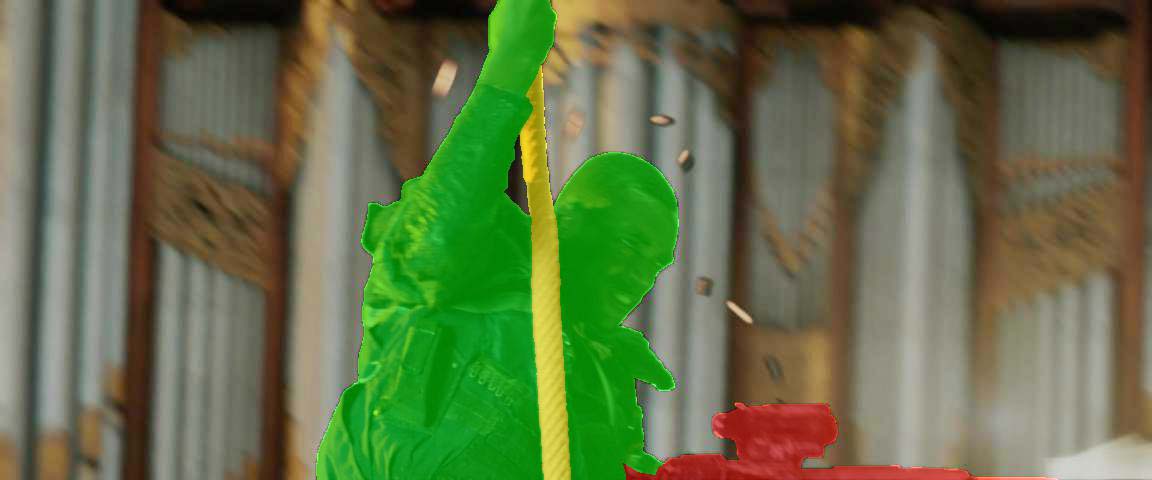} \\
&& \\
\includegraphics[width=0.32\textwidth]{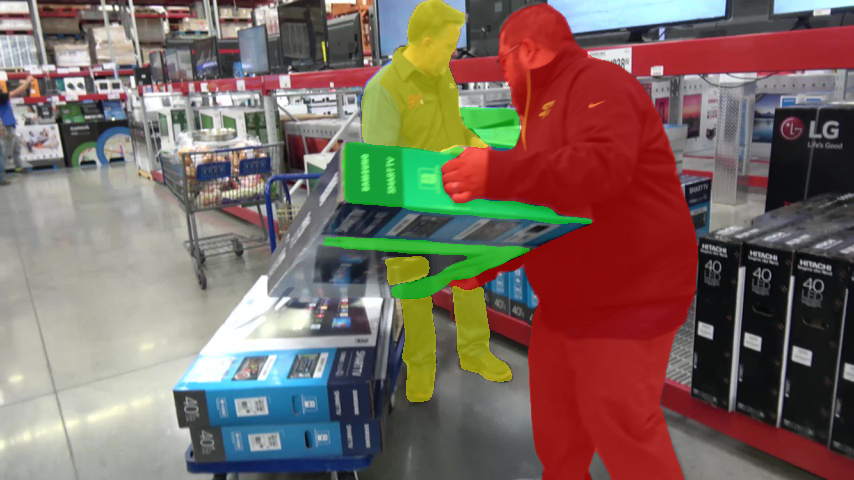} &
\includegraphics[width=0.32\textwidth]{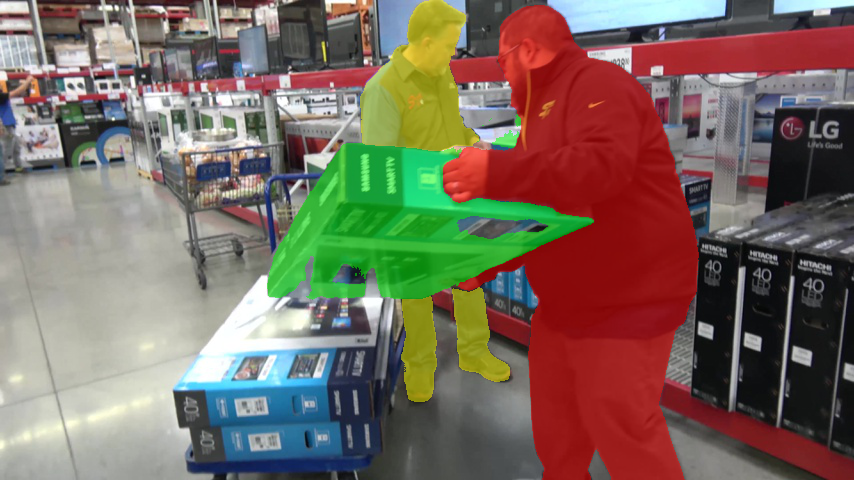} &
\includegraphics[width=0.32\textwidth]{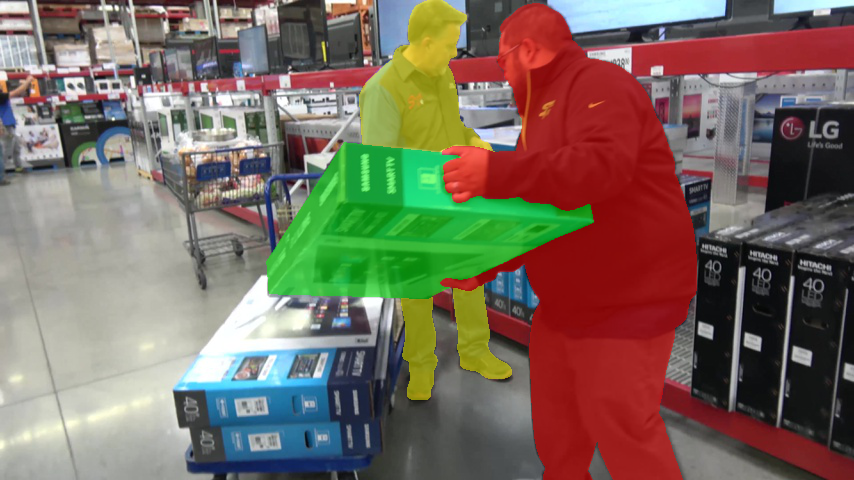} \\
&& \\
\includegraphics[width=0.32\textwidth,trim=400 150 0 20,clip]{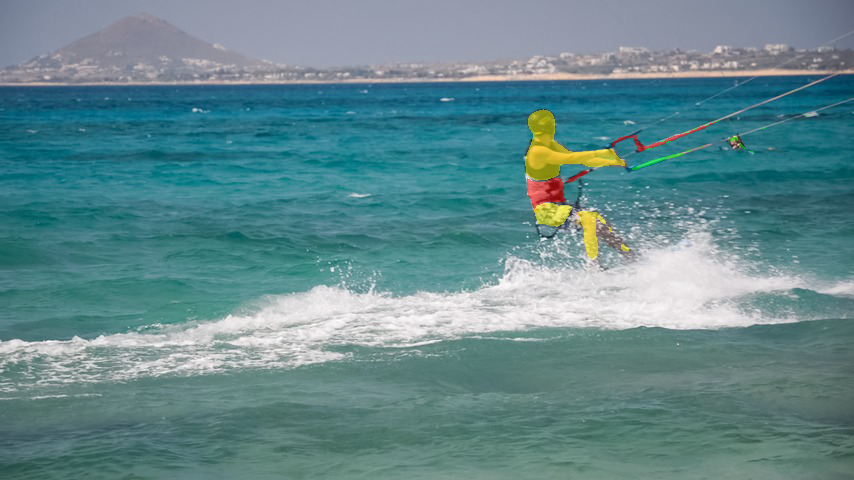} &
\includegraphics[width=0.32\textwidth,trim=400 150 0 20,clip]{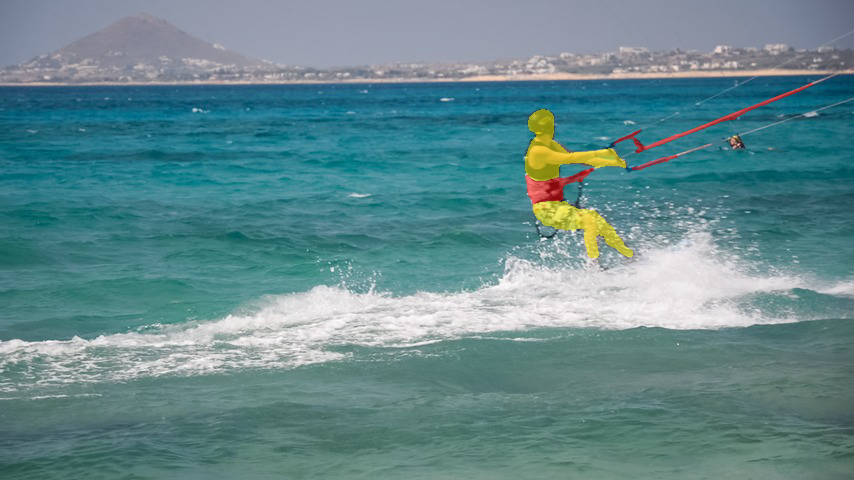} &
\includegraphics[width=0.32\textwidth,trim=400 150 0 20,clip]{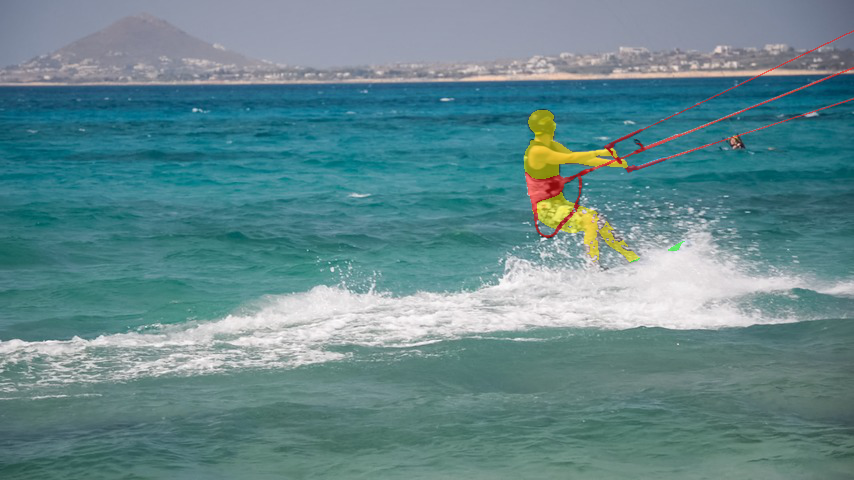} \\
\end{tabular}%
\vspace{0.5mm}%
\caption{Qualitative results on the DAVIS \textit{val} dataset. In the top row, the baseline confuses the background for the gun due to visual similarity, while our approach is successful. In the middle row, our approach produces a more complete segmentation of the green box, and achieves good separation between the background and the yellow leg. In the bottom row, our approach produces better segmentation for fine objects due to the use of high-resolution features.}
\label{fig:qualitative_results1}
\vspace{-2mm}
\end{figure*}

Figure \ref{fig:qualitative_results1} shows some qualitative results comparing our approach to the LWL baseline on the DAVIS 2017 \textit{val} dataset.
The top row shows an example for how our approach is resilient against distractors in the background.
While the baseline confuses an element in background with the gun since they have similar textures, our approach performs a more robust discrimination.
On the second row, our approach produces a more accurate segmentation of all objects without bleeding to other intersecting objects.
On the final row, the impact of incorporating high-resolution features is evident as our approach successfully detects fine objects.

\section{Additional Implementation Details}

In addition to adding a channel for the input and output distractors, we keep the architectures of the mask-encoder and few-shot learner fundamentally unchanged.
However, the number of channels in the encoded mask representations is increased from 16 to 32 in the best model, to accommodate the distractors.
In addition, it proved necessary to configure the few-shot learner to use five rather than three iterations of its internal optimizer, for our model to reach maximum performance.
On the other hand, increasing the number of iterations for the baseline yielded no improvements.

\end{document}

%% file: 00-setup.tex
\newcommand{\mcJ}{\mathcal{J}}  
\newcommand{\mcF}{\mathcal{F}}  
\newcommand{\mcG}{\mathcal{G}}

\newcommand{\mcJu}{\mathcal{J}_\mathrm{u}} 
\newcommand{\mcFu}{\mathcal{F}_\mathrm{u}} 
\newcommand{\mcGu}{\mathcal{G}_\mathrm{u}}

\newcommand{\mcJs}{\mathcal{J}_\mathrm{s}} 
\newcommand{\mcFs}{\mathcal{F}_\mathrm{s}} 
\newcommand{\mcGs}{\mathcal{G}_\mathrm{s}}

\newcommand{\mcJF}{\mathcal{J \& F}} 
\newcommand{\etal}{\emph{et al.}}

%% file: 20-related_work.tex
\section{Related Work}

The video object segmentation (VOS) task can be tackled in a semi-supervised or an unsupervised manner, but we only consider the former in this paper.
Semi-supervised approaches from the literature can usually be depicted as either generative or discriminative.
Generative approaches \cite{vos_joakim,vos_stm} typically focus on constructing a robust model of the target of interest ignoring other objects in the scene. 
On the contrary, discriminative approaches \cite{vos_cvpr,vos_kmn} attempt to solve the task as a classification problem between the target an the background.
A more recent method \cite{vos_lwl} follow an embedding approach to learn features that are as discriminative as possible. 
With the emergence of deep learning, the robustness of feature representations has significantly improved, boosting the performance of most variants of VOS approaches.
However, the current top-performing semi-supervised VOS methods are mainly discriminative, taking into account both the target and the background when solving the task.
Therefore, we focus on discriminative approaches in this paper.


\textbf{Discriminative Video Object Segmentation}~~
Discriminative VOS methods were introduced quite recently.
Yang \etal~\cite{vos_cfbi} proposed to build a target model from separate target and background feature embeddings, extracted from past images and target masks. 
For test frames, extracted features are matched against the two pretrained models of the target and the background. 
STM \cite{vos_stm} incorporates a feature memory, encoding past images and segmentation masks. 
Similarly, Lu \etal~\cite{vos_graphmem} introduce an episodic graph memory unit to mine newly available features, and update the model accordingly.  
In contrast to Yang \etal, both methods \cite{vos_stm,vos_graphmem} produce the final target mask using a dedicated decoder network, from concatenated memory features and new image features. 
Seong \etal~\cite{vos_kmn} extended STM \cite{vos_stm} further with a soft Gaussian-shaped attention mask to limit confusion with distant objects. 
Robinson \etal~\cite{vos_cvpr} introduced the use of discriminative correlation filters to construct a target model that produces a coarse segmentation mask.
This coarse mask is then enhanced and refined through a decoder network.
Learning-what-to-learn \cite{vos_lwl} improved it further by learning to produce target embeddings that are more reliable for training the target model.
All of these aforementioned papers adopt a one-vs-one classification between the target and the background, where other objects in the sequence are considered as background.
In contrast, our approach reduces the likelihood of predicting false positives when some background objects share visual similarities with the target. 


\textbf{Distractor-Aware Modelling}~~Distractor-aware modeling can be realized as a kind of hard-example mining when training a model.
The general concept is to identify inputs that are more likely to confuse a given model and emphasis them during training, 
One of the earliest uses of this concept is found in the classical human detection algorithm, histograms of oriented gradients~\cite{det_dalal-trigs_hog}. 
A more recent example \cite{det_uhem} proposed marking flickering object detections in a video as hard-negatives, assigning them higher priority during training. 
A recent visual object tracking method \cite{track_kys} ranks target proposals based on their signal strength, where the strongest is assumed to be the target and the rest are distractors. 
The method tracks the complete scene state in a dense vector field over all spatial locations, using it to classify regions as either target, distractor or background. 
A similar Siamese-based approach \cite{track_dasiamrpn} also classifies a target and any distractors through ranking of detection strengths. 
In both approaches, distractors are provided as hard examples during online training their respective tracker target models.
Zhu \etal~\cite{track_dasiamrpn} also introduced distractor awareness to Siamese visual tracking networks as they noticed that the standard trackers posses imbalanced distribution of training data leading to less discriminative features.
They propose an effective sampling strategy to make the model more discriminative towards semantic distractors.
In this paper, we follow the strategy adopted by these approaches, and we introduce distractor-awareness to the task of video object segmentation.






%
%


\textbf{Segmentation Mask Upsampling and Refinement}~~
Existing VOS CNNs employ different upsampling and refinement approaches to provide the final segmentation mask at the full resolution.
RGMP and STM \cite{vos_rgmp,vos_stm} employ two residual blocks with a bilinear interpolation in between for refinement and upsampling.
Seong \etal~\cite{vos_kmn} used a residual block followed by two refinement modules with skip connections. 
In contrast, Robinson \etal and Bhat \etal \cite{vos_cvpr,vos_lwl} replace the residual blocks with standard convolution, and employs bicubic instead of bilinear interpolation.
All these approaches provide spatially independent prediction with no regularization to enforce local consistency, especially at uncertain regions such as edges and object boundaries.
We employ the convex upsampler \cite{raft} to jointly upsample the final mask while enforcing spatial consistency.



%% file: 30-method.tex
\section{Method}

Ideally, in video object segmentation, it is desired to produce pixel-wise predictions $y$ either as \emph{target} $\mathcal{T}$ or \emph{background} $\mathcal{B}$. 
In a probabilistic sense, we are interested in maximizing the posterior probability for the target given an input embedding $X$:

\begin{align} \label{eq:1}
\nonumber P(Y=\mathcal{T} | X) &= \dfrac{ P(X,Y=\mathcal{T}) }{P(X)} 
          = \dfrac{ P(X,Y=\mathcal{T}) }{ P(X,Y=\mathcal{T}) + P(X,Y=\mathcal{B}) } \\
         &= \dfrac{ 1 }{ 1 + \dfrac{P(X,Y=\mathcal{B})}{P(X,Y=\mathcal{T})} }
          = \dfrac{ 1 } { 1+ \dfrac{ P(X | Y=\mathcal{B}) P(Y=\mathcal{B}) }{ P(X | Y=\mathcal{T}) P(Y=\mathcal{T})} } \enspace ,
\end{align}

The ratio in the denominator determines the posterior probability. 
If a pixel belongs to the target, the ratio becomes small and the posterior probability tends to 1.
Contrarily, if a pixel belongs to the background and is quite distinct from the target, the ratio becomes large and the posterior probability goes to 0.
However, if the target prior is large, and the background and target likelihoods are similar because $X$ contains features of a distractor, the posterior can easily be larger than 0.5 and produce false positives. 
We propose splitting the non-target into two classes, background $\mathcal{B}$ and distractor $\mathcal{D}$:

\begin{align} \label{eq:2}
\nonumber P(X, Y \neq \mathcal{T}) &= P(X, Y = \mathcal{B}) + P(X, Y = \mathcal{D}) \\ 
                                   &=  P(X | Y=\mathcal{B}) P(Y=\mathcal{B}) + P(X | Y=\mathcal{D}) P(Y=\mathcal{D}) \enspace .
\end{align}  

Consequently, the ratio in (\ref{eq:1}) will have two terms in the enumerator as denoted by (\ref{eq:2}).
This modification limits the occurrences of false positives as it models ambiguous pixels from the first frame and propagates them.
As an example, if the likelihood of a certain pixel is similar between the target and the distractor at an intermediate frame, the propagated prior from previous frames will cause the ratio to be large, and the probability to drop.
In the following sections, we will describe how to modify an existing baseline to be distractor-aware.

\begin{figure*}[ht]
\includegraphics[width=1.0\textwidth]{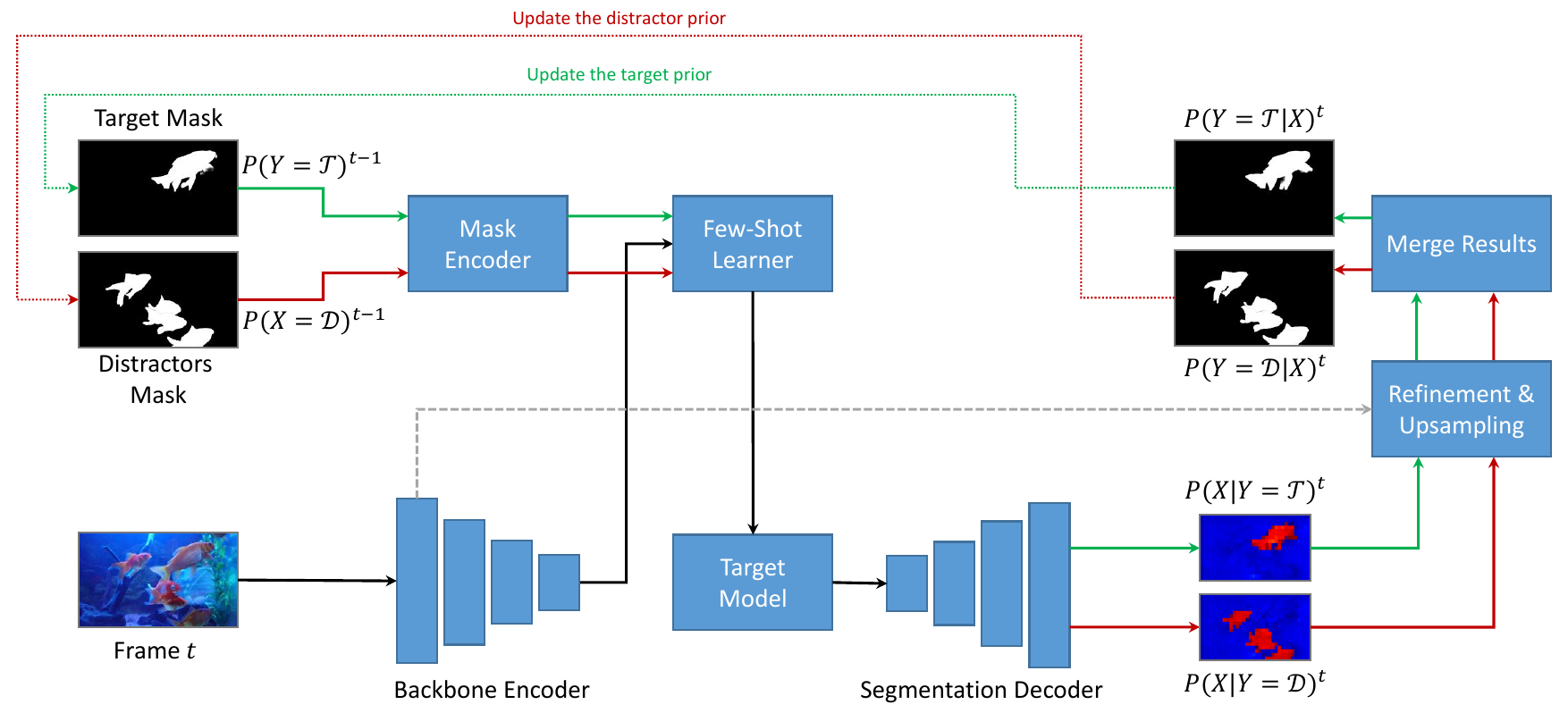}
\caption{An overview of our distractor-aware approach. We extend the baseline LWL to incorporate information about distractors throughout its mask encoder, target model, and segmentation decoder stages, and replace its upsampler with a joint refinement and upsampling approach, acting as local regularization.}
\label{fig:overview}
\end{figure*}


\subsection{Baseline Approach}

We base our approach on the recently published method learning-what-to-learn \cite{vos_lwl} (LWL).
In the LWL baseline, a target segmentation mask is encoded by a mask encoder into an multi-channel embedding at one sixteenth of the original image resolution. 
At the same time, a standard backbone encoder network (ResNet-50) is used to extract features from the whole image (see Figure \ref{fig:overview}). 
At the first frame, both the mask embeddings and the image features are utilized as training data for a target model $T_\theta(x) = x \ast \theta$ that is trained with a few-shot learner.
In subsequent video frames, this target model generates new multi-channel embeddings from the corresponding image features. 
A decoder network finally recovers segmentation masks at full resolution from these embeddings.
Eventually, the newly predicted masks and deep features are added to the training data, so that the target model can be tuned, adapting it to the changing target appearance.
Note that multiple objects are tracked independently and individual target models do not interact with each other.
In other words, the target model is not aware of any objects in the scene since everything other than the target is treated as background.




\subsection{Introducing Distractor-Awareness}


As describe in the related work section, there are several ways to detect distractors. 
Here, we consider other objects in the scene as potential distractors.
More specifically, under the assumption that segmentation masks are binary, given a target masks $\mathbf{1}_{t_i}$, we generate a distractor mask as: 
\begin{equation} \label{eq:dist}
    \mathbf{1}_{d_i} = \underset{j \neq i}{\cup} \mathbf{1}_{t_j} \  \forall j \in \mathbb{I} \enspace,
\end{equation}
where $\mathbb{I} = \{1 ... N \}$ is the set of target IDs in the current video sequence. 
Both the target and the distractor masks are set to the ground truth in the first frame, and updated with the previous predictions in later frames.
To accommodate the distractor, we add a second input channel to the mask encoder and a second output channel to the  segmentation decoder (see figure \ref{fig:overview}). 


As the segmentation of frames progresses, we merge the decoded and upsampled target masks to form new distractors.
However, in this case, the propagated masks are no longer binary and (\ref{eq:dist}) needs to be replaced. For this, we develop a per-pixel winner-take-all (WTA) function.

Let $p_{t_i}(x) \in \mathbb{R}^{H \times W}$ be the target segment probability map, i.e. the network decoder output after a sigmoid activation, of the target with index $i$.
Now let
\begin{equation} \label{eq:dists}
p_{\max}(x) = \underset{j}{\sup}\, p_{t_j}(x) \  \forall j \in \mathbb{I} \enspace,
\end{equation}

\begin{equation} \label{eq:disti}
    p_{\min}(x) = \underset{j}{\inf}\, p_{t_j}(x) \  \forall j \in \mathbb{I} \enspace,
\end{equation}
merging the highest and lowest probabilities (per pixel), into $p_{\max}$ and $p_{\min}$.

Now let, $L(x) \in (\mathbb{I} \cup \{0\})^{H \times W}$ be the map of merged segmentation labels (after softmax-aggregation, as introduced in \cite{vos_rgmp}), with 0 the background label.

Also let
\begin{equation}
    \mathbf{1}_{f}(x) = {
    \begin{cases}
        1~&{\text{ if }}~L(x) > 0~, \\
        0~&{\text{ if }}~L(x) = 0~.
    \end{cases}}
\end{equation}
indicate regions with any foreground pixel, and
\begin{equation}
\mathbf{1}_{d_i}(x) = {
\begin{cases}
    \mathbf{1}_{f}(x)~&{\text{if}}~L(x) \neq i~, \\
    0~&\text{otherwise}~.
\end{cases}}
\end{equation}
indicate distractors of target $i$. A new probability map of distractor $i$ is generated as
\begin{equation}
p_{d_i}(x) = \mathbf{1}_{d_i}(x) p_{\max}(x) + (1 - \mathbf{1}_{f}(x)) p_{\min}(x).
\end{equation}
In less formal terms, we let the the most certain predictions of target and background \say{win} in every pixel. $p_{d_i}$ takes information from every $p_{t_j}$ except that of target $i$.

\begin{figure*}[ht]
\includegraphics[width=1.0\textwidth]{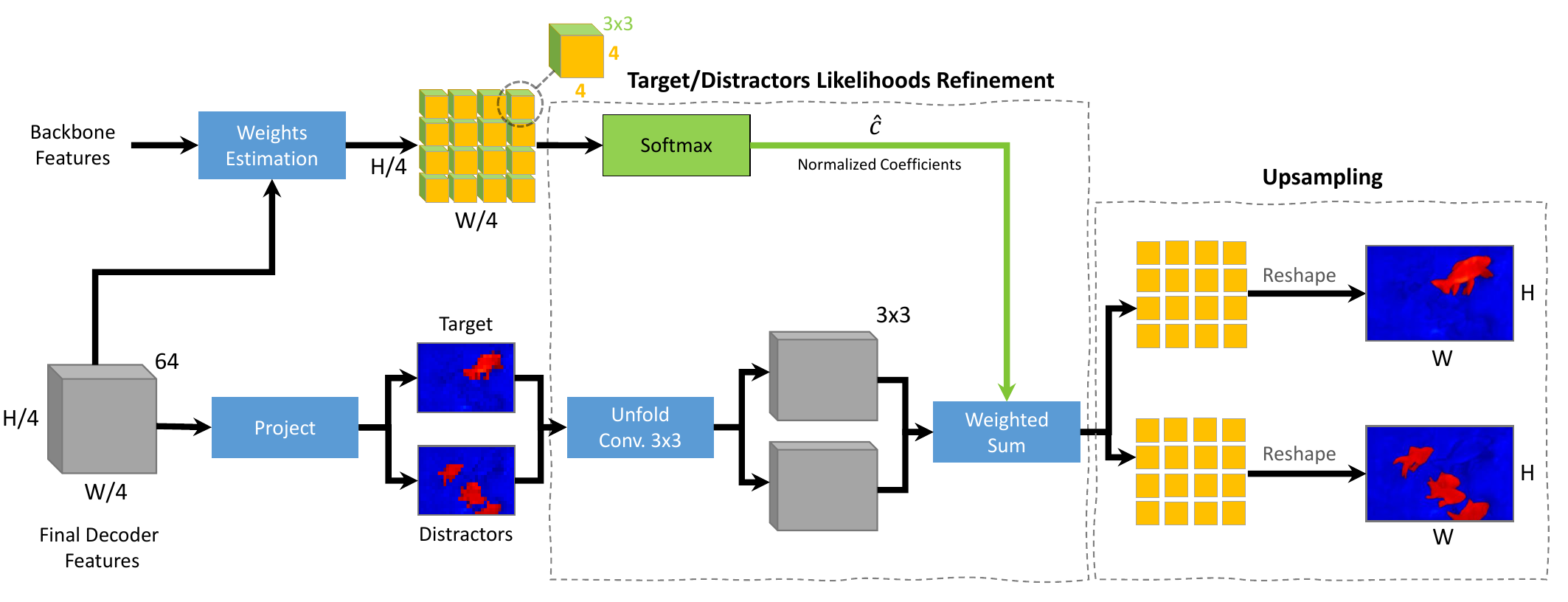}
\caption{An overview of the refinement and upsampling module. The final decoder features are first projected into target and distractor logit maps. A weights estimation network then predicts a 5D tensor of weights in $3 \times 3$ windows around each pixel in the logit maps (shown in green), as well as interpolation data (shown in yellow). The weights are mapped to normalized probability vectors with the softmax function and then used refine the target and the distractor logits with weighted summation. The refined logits are finally upsampled to full resolution with the interpolation data.}
\label{fig:upsampler-diag}
\end{figure*}

In the ablation study, we compare the performance of this function to that of feeding back the decoded distractor to the few-shot learner without modification.

\subsection{Joint Refinement and Upsampling for Local Consistency} \label{sec:up}

The majority of existing VOS methods adopt a binary classification scheme between the foreground and the background.
It is typically challenging to produce accurate predictions at uncertain regions such as target boundaries due to multi-modality along edges.
This problem is aggravated when we introduce a new class for distractors, as the decision is now made between three classes instead of two.
We tackle this problem by introducing two modifications to the baseline encoder and decoder, respectively.

First, we provide high-resolution feature maps from the backbone when training the target model in the few-shot learner. Unlike the baseline, we employ backbone features with a 1/8th of the full resolution instead of 1/16th. These higher-resolution features provide finer details to the target model, especially along edges, to help resolving ambiguities in uncertain regions.
Second, we replace the baseline upsampler on the decoder output, with a joint refinement and upsampling unit based on the convex combination module of \cite{raft}.
This is originally employed to upsample flow fields, while we adopt it to produce consistent selections between the target and distractors in logit space.

The proposed joint refinement/upsampling approach is illustrated in Figure \ref{fig:upsampler-diag}. 
First, the output from the decoder is projected using two convolution layers to two channels resembling the likelihoods of the target and the distractor.
Then, the likelihoods are unfolded into $3\times3$ patches around each pixel for computational efficiency.
In parallel, we employ a weights estimation network that jointly perform likelihoods refinement, and mask upsampling. The former allows modifying the likelihoods of the target and the distractor based on their neighbors to enforce some local consistency, while the latter produces pixel values needed for upsampling.


For the the refinement, the weights estimation network predicts a local coefficients vector $\mathbf{c}_x$ for a $3\times3$ window centered around each pixel $x \in X$:
\begin{equation}
    \mathbf{c}_x = [c_{-4}, \dots,  c_{-1} c_{-1}, c_{0}, c_{1}, \dots, c_{4}] \enspace,
\end{equation}
\noindent where $c_0$ is the coefficient on top of $x$.
However, these coefficients are not normalized, and to preserve likelihood ratios we first map $\mathbf{c}_x$ to a normalized vector $\hat{\mathbf{c}}_x$ using the softmax function. 
To modify the likelihoods, we subsequently apply the coefficients to the likelihoods using a weighted sum resembling convolution:
\begin{equation}
    P(X|Y)' * \hat{\mathbf{c}}_x [x] = \sum_{m \in [-4:4]} P(X|Y)[x-m] \ \hat{\mathbf{c}}_x[m] \enspace.
\end{equation}

For the upsampling, we need to predict pixel values for $4\times4$ patches around each pixel for a scaling of 4.
Those values are also predicted using the weights-estimation network and are needed to produce the full resolution prediction.
This combined refinement-and-upsampling feature volume takes the shape of a local 3D tensor $\mathbf{A}_x \in \mathbb{R}^{4\times4\times9}$ for each pixel $x \in X$.


%% file: 40-experiments.tex
\section{Experiments}

\newcommand{\yes}{\checkmark}
\newcommand{\no}{\text{\ding{55}}}
\newcommand{\red}[1]{\textbf{\textcolor{red}{#1}}}
\newcommand{\blue}[1]{\textbf{\textcolor{blue}{#1}}}
\newcommand{\black}[1]{\textbf{\textcolor{black}{#1}}}

In this section, we provide implementation details including the training procedure and loss function.
We compare against state-of-the-art approaches on DAVIS 2017 \cite{dv2017}, and YouTube-VOS 2018 \cite{yt2018}.
In addition, we provide an extensive ablation analysis.

\subsection{Implementation Details}
\label{sec:implementation}

\subsubsection{Training procedure}

Similar to the baseline, our training setup imitates the segmentation processing during inference. 
Each training sample is a mini-video of four frames with one main target object to segment. 
The few-shot learner is provided the first frame to train a target model. 
The target model and the decoder then predicts segmentation masks from the subsequent three frames. 
The predictions in turn, are both used to update the target model and compute the network training loss. 

To extend this procedure for distractors, we replicate the segmentation process for all other objects that are present in the ground-truth label map.
These additional segments are then merged and provided as distractors to the main target. 
However, we set the network into evaluation mode when processing these targets to conserve memory.

We employ the same data augmentation and first two training phases as the baseline \cite{vos_lwl}.
The proposed refinement and upsampling module is trained in a third phase for 6,000 iterations on DAVIS 2017, with learning rate $10^{-4}$, all other weights frozen.

\subsubsection{Loss Function}

LWL was trained with the Lovász-softmax loss \cite{lovasz_loss}, a differentiable relaxation of the Jaccard similarity measure, and we continue to do so here. However, in the SOTA experiments, we did initially not see any improvements with our method on YouTube-VOS. We hypothesize that this is due to large size-differences between objects in the same sequence in YouTube-VOS. To counter this, we split the loss into two terms, or a balanced loss: 
\begin{equation}
L = \mathrm{Lovasz}(T) + w(\hat{T},\hat{D})\mathrm{Lovasz}(D)
\end{equation}
where $T$ and $D$ are the output batch from the decoder network, separated into target and distractor channels. $w(\hat{T},\hat{D})$ reduces the influence of large objects as a function of the number of ground-truth target pixels $|\hat{T}|$ and ground-truth distractor pixels $|\hat{D}|$ across the training batch:
\begin{equation}
    w(\hat{T},\hat{D}) = \min(|\hat{T}|/|\hat{D}|, 1.0)
\end{equation}
In other words, when the distractors jointly occupy a larger region than the target, their influence on the loss is reduced.

\subsubsection{Relaxed distractor loss}

Some sequences have only one target. This implies that no distractors exist, since we derive them from every other target.
To not unnecessarily over-constrain the training, we partially disable the computation of the loss in training samples in these cases.
Specifically, we require the distractor output to be zero in the area under the target, but allow it to take any value elsewhere.

We also train a variant with a \say{hard} loss, requiring that no distractor is output when no distractor is given, and compare them in the experiments.


\subsection{State-of-the-art Comparisons}

\begin{table}
\centering
\input{tbl_dv_yt}
\vspace{1mm}
\caption{Results on DAVIS 2017 and YouTube-VOS 2018, comparing our method to the LWL baseline and the state-of-the-art. The LWL scores are from our own run with the official code.}
\label{tab:dv_yt}
\end{table}

We compare against the most recent state-of-the-art approaches for video object segmentation: CFBI \cite{vos_cfbi}, STM \cite{vos_stm}, KMN \cite{vos_kmn}, GMVOS \cite{vos_graphmem}, and  the baseline LWL \cite{vos_lwl}. CFBI-MS is a multi-scale variant operating on three different scales. Our method in contrast, is single scale and thus more comparable to the plain CFBI. Running the official LWL implementation we noticed that our results do not coincide exactly with those reported in \cite{vos_lwl}.
Therefore, we use these newly obtained scores to accurately compare the baseline to our method.

Table \ref{tab:dv_yt} summarizes the quantiative results on DAVIS 2017 and the YouTube-VOS 2018 validation split.
On DAVIS val, our proposed approach without the balanced loss sets a new state-of-the-art on DAVIS val, while improving over the baseline on test-dev by 3.6 percentage points. With the balanced loss however, test-dev surprisingly improves 4.6 percentage points, while lowering the gain on val to 2.5 percentage points.
On the YouTube-VOS 2018 validation split, our approach improves the baseline by 0.8 percentage points, when trained \textit{with} the balanced loss, while scoring similarly to other state-of-the-art approaches.

Some qualitative results are found in the supplement.


\subsection{Ablation Study}

\label{sec:ablation}

\begin{table}
\centering
\input{tbl_ablation}
\vspace{1mm}
\caption{Ablation study on DAVIS 2017.  See Section \ref{sec:ablation} for details. Interesting scores are printed in bold type.
} 
\label{tab:ablation}
\end{table}

In this section, we analyze the impact of different components of our method on the overall performance.
We use the DAVIS 2017 dataset \cite{dv2017} for this purpose, where we include both \textit{val} and \textit{test-dev} splits for diversity.
The results are shown in Table \ref{tab:ablation}.
Since LWL \cite{vos_lwl} is our baseline method, we reevaluate their official pretrained model.
As mentioned above, there is a discrepancy between our obtained scores and the published ones.
This could be due to several factors, e.g. hardware/software differences and driver variations.
However, we use the scores that we obtained for a valid comparison.

We first enable or disable various combinations of these components: the high-resolution features (L2), the distractor-awareness (D), the refinement/upsampling module (U).
Enabling D alone causes a slight improvement. Adding U to this improves the test-dev split a further.
We believe that the lack of high-resolution features prevents further gains.
Similarly, introducing L2 alone, causes a marginal improvement on both splits and enabling U at the same time, shows no improvement on the valdiation set, and hurts the performance on the test-set.
This can be explained by our argument in section \ref{sec:up}: When the decision is made between two classes (target and background), there is no need for the upsampler to resolve ambiguities.
As to why this combinatios hurts test-dev performance, is unclear.
However, enabling all three (L2+D+U) at the same time, yields significantly better results over the baseline.

With L2, D and U included, we then separately replace the relaxed distractor loss with the \say{hard} loss (Hard loss). The results then drops back to the baseline for DAVIS validation split, but oddly \textit{improves} on the test-dev split.

We also test a pair of inference-time modifications. First, we zero out the distractor masks so that the few shot learner do not use them. Second, we replace the WTA distractor-merging function with a simple pass-through; Distractors are directly routed from the decoder output back to the few-shot learner, bypassing the merging step. As one would expect, the results drop in both cases, proving that WTA is important. Surprisingly though, disabling the WTA function hits DAVIS test-dev split much harder than the validation split.

Finally, we add the balanced loss (B) to the L2+D+U variant. This clearly damages the DAVIS validation scores but improves the test-dev results.
However, at the same time it also improves the YouTube-VOS results in Table \ref{tab:dv_yt}.

These results suggest that the method works, but also reveal differences between the DAVIS dataset splits that we cannot explain at this time.


\subsection{A Note on WTA vs. Softmax Aggregation}

The softmax aggregation introduced in \cite{vos_rgmp}, and used to merge estimations of multiple targets, was introduced as a superior alternative to a winner-take-all approach. In our case however, we found that to merge distractors, WTA performs better. We hypothesize that lacking a dedicated background channel is beneficial to the refinement and upsampling module, as the decoder may output low activations on both target and distractors when the classification is uncertain.
 


\subsection{Emergent distractors}

\begin{figure}[ht!]
\begin{tabular}{ccccc}
    RGB Image & (R) Target & (R) Distractor & (H) Target & (H) Distractor \\  
    \includegraphics[width=0.19\columnwidth]{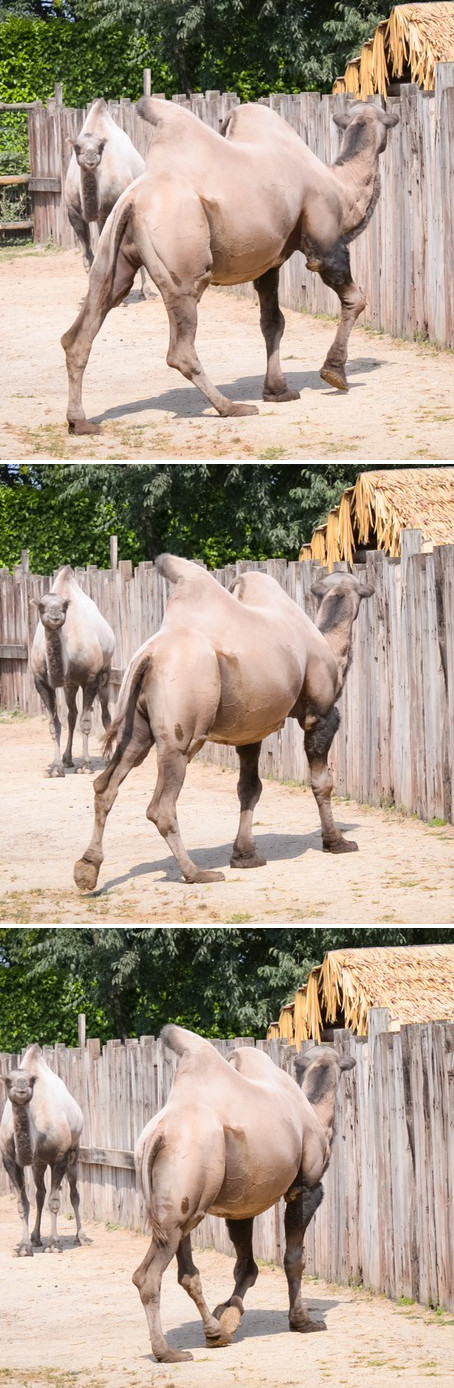} &
    \includegraphics[width=0.19\columnwidth]{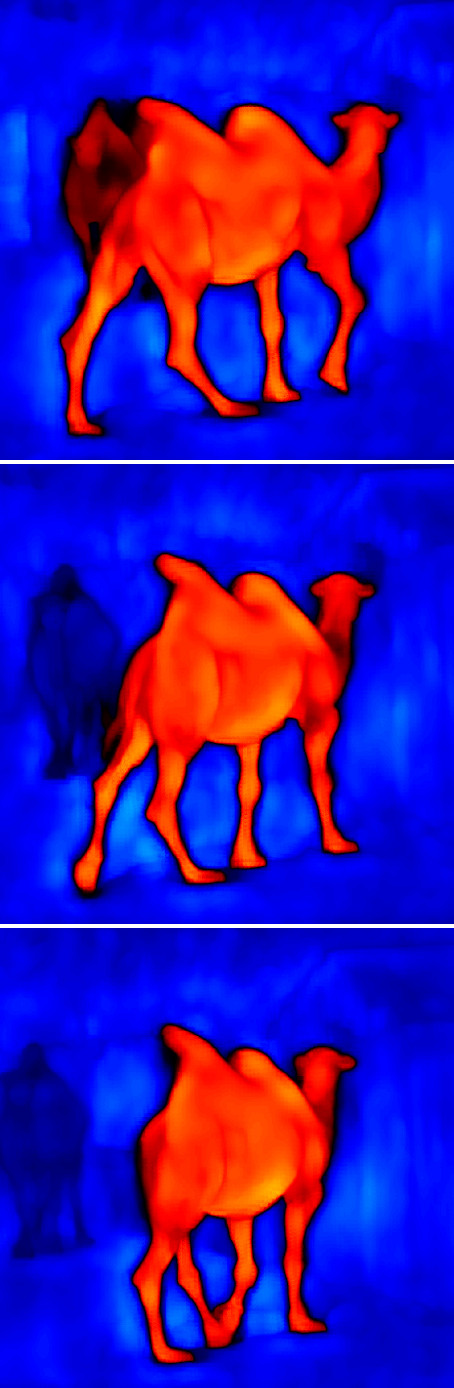} &
    \includegraphics[width=0.19\columnwidth]{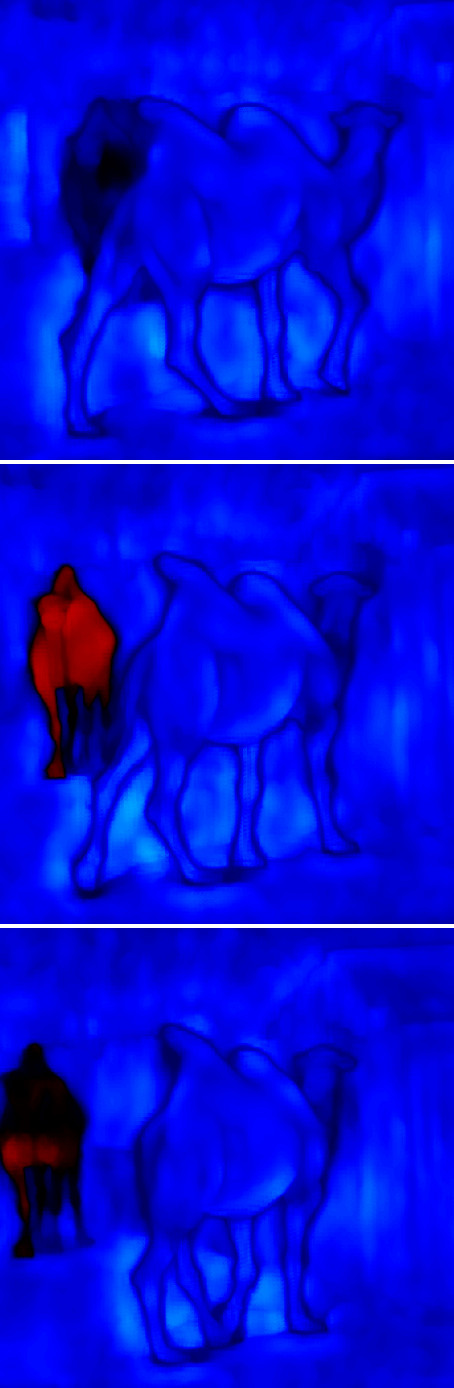} &
    \includegraphics[width=0.19\columnwidth]{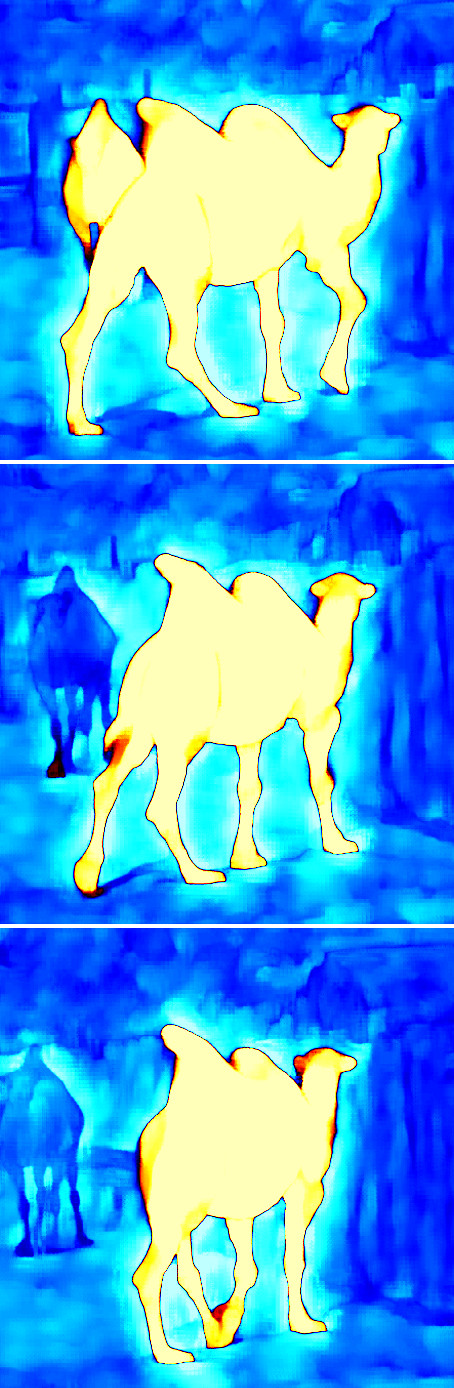} &
    \includegraphics[width=0.19\columnwidth]{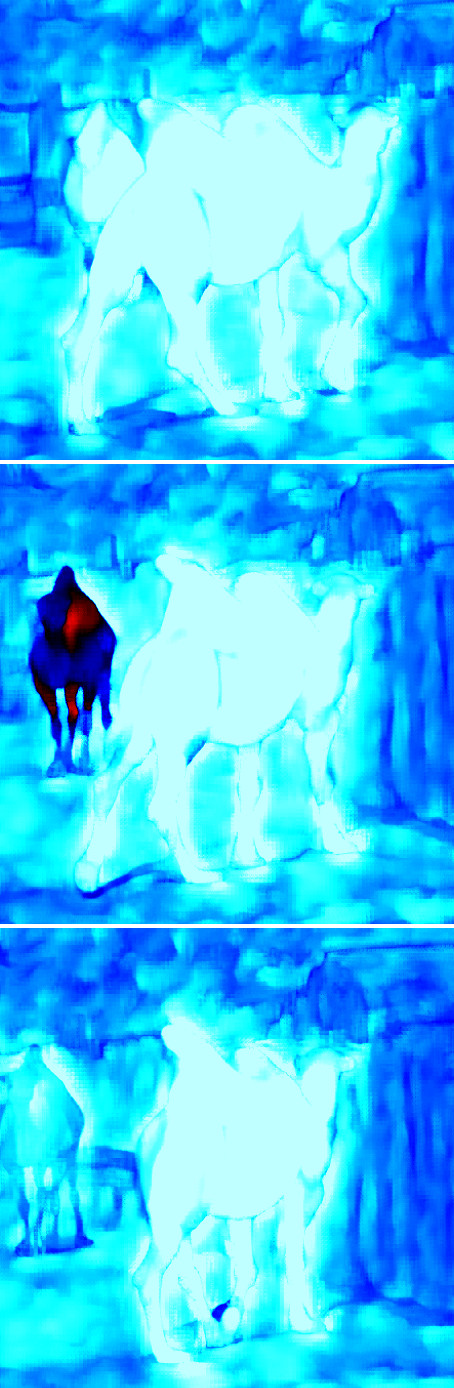} //
\end{tabular}
\caption{Frames 40, 50 60 (from top to bottom) from the 'camel' sequence, with target/distractor score maps. Our model can learn to identify distractors in case they were not explicitly provided, provided it is trained with the relaxed distractor loss. Red/yellow indicates positive log-likelihoods, blue/cyan negative. The colors represent the same value range in all columns. (R) and (H) indicate results from models trained with the relaxed and hard distractor losses, respectively.}
\label{fig:camel}
\end{figure}

An essential question is how our framework would behave when there is only one labeled object in the scene.
Interestingly, the model learns to identify ambiguous regions.
Figure \ref{fig:camel} shows an example where our approach learns to identify the camel in the background without any explicit supervision (if columns 2+3 from the left). We attribute this to the relaxed distractor loss described in Section \ref{sec:implementation}. To test this, we modify the loss to be \say{hard} (H). As suspected, this suppresses the behaviour greatly, while increasing the decoders' certainty in both the target and background (columns 4+5).

\setlength{\tabcolsep}{0.5pt}
\setlength{\tabcolsep}{6pt}


%% file: tbl_dv_yt.tex
{
\newcommand{\bb}[1]{\textbf{#1}}
\begin{tabular}{l|ccc|ccc|ccccc}
\toprule
                Method & \multicolumn{3}{c|}{DAVIS'17 val}             & \multicolumn{3}{c|}{DAVIS'17 test-dev} &   \multicolumn{5}{c}{YouTube-VOS '18 valid}                \\
                  Name &                    $\mcG$ &    $\mcJ$ &    $\mcF$ &    $\mcG$ &    $\mcJ$ &    $\mcF$ &   $\mcG$ & $\mcJs$ & $\mcJu$ & $\mcFs$ & $\mcFu$           \\
\midrule
                  CFBI \cite{vos_cfbi} &      81.9 &      79.1 &      84.6 &      74.8 &      71.1 &      78.5 &       81.4 &      81.1 &      75.3 &      85.8 &      83.4 \\
               CFBI-MS \cite{vos_cfbi} & \bb{83.3} & \bb{80.5} & \bb{86.0} & \bb{77.5} &      73.8 & \bb{81.1} &  \bb{82.7} & \bb{82.2} & \bb{76.9} & \bb{86.8} & \bb{85.0} \\
                   KMN \cite{vos_kmn}  &      82.8 &      80.0 &      85.6 &      77.2 & \bb{74.1} &      80.3 &       81.4 &      81.4 &      75.3 &      85.6 &      83.3 \\
                   STM \cite{vos_stm}  &      81.8 &      79.2 &      84.3 &      72.2 &      69.3 &      75.2 &       79.4 &      79.7 &      72.8 &      84.2 &      80.9 \\
                 GMVOS \cite{vos_graphmem} &  82.8 &      80.2 &      85.2 &        -  &        -  &        -  &       80.2 &      80.7 &      74.0 &      85.1 &      80.9 \\
 \hline
                     LWL \cite{vos_lwl}&      80.1 &      77.4 &      82.8 &      70.8 &      68.0 &      73.7 &       80.7 &      79.5 &      75.6 &      84.0 &      83.5 \\
                    \bb{Ours}          & \bb{83.7} & \bb{81.1} & \bb{86.2} &      74.1 &      71.2 &      77.1 &       79.8 &      79.8 &      74.0 &      84.1 &      81.3 \\
                    \bb{Ours} (Balanced loss) & 82.6 &    80.8 &      85.3 & \bb{75.4} & \bb{72.5} & \bb{78.2} &  \bb{81.5} & \bb{80.4} & \bb{76.0} & \bb{85.1} & \bb{84.5} \\
\bottomrule
\end{tabular}
}

%% file: tbl_ablation.tex
\newcommand{\bb}[1]{\textbf{#1}}
\begin{tabular}{cccc|ccc|ccc}
\toprule
\multicolumn{4}{c|}{Parameters}        & \multicolumn{3}{c|}{DAVIS'17 val} & \multicolumn{3}{c}{DAVIS'17 test-dev} \\
          L2 &     D &     U &       B &          $\mcG$ & $\mcJ$ & $\mcF$ &            $\mcG$ & $\mcJ$ & $\mcF$ \\
\midrule
             &       &       &         &            80.1 &   77.4 &   82.8 &              70.8 &   68.0 &   73.7 \\
             &  \yes &       &         &            80.7 &   77.9 &   83.4 &              71.0 &   67.7 &   74.2 \\
             &  \yes &  \yes &         &            80.7 &   78.2 &   83.3 &              71.7 &   68.7 &   74.8 \\
        \yes &       &       &         &            81.9 &   78.9 &   84.9 &              72.7 &   69.5 &   75.9 \\
        \yes &       &  \yes &         &            81.6 &   78.9 &   84.3 &              71.6 &   68.5 &   74.7 \\
        \yes &  \yes &       &         &            81.3 &   78.8 &   83.9 &              73.2 &   70.0 &   76.4 \\
        \yes &  \yes &  \yes &         &       \bb{83.7} &   81.1 &   86.2 &         \bb{74.1} &   71.2 &   77.1 \\
        
\midrule

\multicolumn{4}{c|}{Hard loss}         &       \bb{80.2} &   77.5 &   82.8 &         \bb{74.9} &   72.4 &   77.4 \\
 \multicolumn{4}{c|}{No Distractors}   &            81.1 &   78.7 &   83.4 &              71.2 &   68.0 &   74.4 \\
 \multicolumn{4}{c|}{No WTA}           &            83.0 &   80.5 &   85.5 &              71.8 &   68.4 &   75.2 \\
\midrule
           \yes &  \yes &  \yes & \yes &       \bb{82.6} &   80.8 &   85.3 &         \bb{75.4} & 72.5 &   78.2 \\
\bottomrule
\end{tabular}

%% file: egbib.bib
@article{dv2017,
    author = {Jordi Pont-Tuset and Federico Perazzi and Sergi Caelles and Pablo Arbel\'aez and Alexander Sorkine-Hornung and Luc {Van Gool}},
    title = {The 2017 DAVIS Challenge on Video Object Segmentation},
    journal = {arXiv:1704.00675}, year = {2017}
}

@article{yt2018,
    title={Youtube-vos: A large-scale video object segmentation benchmark},
    author={Xu, Ning and Yang, Linjie and Fan, Yuchen and Yue, Dingcheng and Liang, Yuchen and Yang, Jianchao and Huang, Thomas},
    journal={arXiv preprint arXiv:1809.03327}, year={2018}
}

@inproceedings{vos_stm,
	title={Video object segmentation using space-time memory networks},
	author={Oh, Seoung Wug and Lee, Joon-Young and Xu, Ning and Kim, Seon Joo},
    booktitle = {2019 IEEE/CVF Conference on Computer Vision and Pattern Recognition (CVPR)}, month={June}, year={2019},
}

@inproceedings{vos_lwl,
    author={Bhat, Goutam and Lawin, Felix J{\"a}remo and Danelljan, Martin and Robinson, Andreas and Felsberg, Michael and Van Gool, Luc and Timofte, Radu},
    title={Learning What to Learn for Video Object Segmentation},
    booktitle = {The European Conference on Computer Vision (ECCV)}, month={August}, year={2020},
}

@InProceedings{vos_cvpr,
    author = {Robinson, Andreas and Lawin, Felix Jaremo and Danelljan, Martin and Khan, Fahad Shahbaz and Felsberg, Michael},
    title = {Learning Fast and Robust Target Models for Video Object Segmentation},
    booktitle = {2020 IEEE/CVF Conference on Computer Vision and Pattern Recognition (CVPR)}, month={June}, year={2020},
}

@inproceedings{vos_cfbi,
    author={Yang, Zongxin and Wei, Yunchao and Yang, Yi},
    title={Collaborative Video Object Segmentation by Foreground-Background Integration},
    booktitle = {The European Conference on Computer Vision (ECCV)}, month={August}, year={2020}, publisher={Springer}
}

@article{vos_kmn,
    title={Kernelized memory network for video object segmentation},
    author={Seong, Hongje and Hyun, Junhyuk and Kim, Euntai},
    booktitle = {The European Conference on Computer Vision (ECCV)}, month={August}, year={2020},
}

@inproceedings{vos_graphmem,  
    title={Video Object Segmentation with Episodic Graph Memory Networks},  
    author={Lu, Xiankai and Wang, Wenguan and Martin, Danelljan and Zhou, Tianfei and Shen, Jianbing and Luc, Van Gool},  
    booktitle = {The European Conference on Computer Vision (ECCV)}, month={August}, year={2020},
}

@inproceedings{vos_joakim,
  title={A generative appearance model for end-to-end video object segmentation},
  author={Johnander, Joakim and Danelljan, Martin and Brissman, Emil and Khan, Fahad Shahbaz and Felsberg, Michael},
  booktitle = {2019 IEEE/CVF Conference on Computer Vision and Pattern Recognition (CVPR)}, month={June}, year={2019},
}

@inproceedings{vos_rgmp,
	title={Fast video object segmentation by reference-guided mask propagation},
	author={Oh, Seoung Wug and Lee, Joon-Young and Sunkavalli, Kalyan and Kim, Seon Joo},
	booktitle = {2018 IEEE/CVF Conference on Computer Vision and Pattern Recognition (CVPR)}, month={June}, year={2018}
}

@inproceedings{det_dalal-trigs_hog,
    title={Histograms of oriented gradients for human detection},
    author={Dalal, Navneet and Triggs, Bill},
    booktitle={2005 IEEE computer society conference on computer vision and pattern recognition (CVPR'05)},
}

@inproceedings{det_uhem,
    author = {Jin, SouYoung and RoyChowdhury, Aruni and Jiang, Huaizu and Singh, Ashish and Prasad, Aditya and Chakraborty, Deep and Learned-Miller, Erik},
    title = {Unsupervised Hard Example Mining from Videos for Improved Object Detection},
    booktitle = {The European Conference on Computer Vision (ECCV)}, month = {September}, year = {2018}
}

@inproceedings{track_dasiamrpn,
    title={Distractor-aware siamese networks for visual object tracking},
    author={Zhu, Zheng and Wang, Qiang and Li, Bo and Wu, Wei and Yan, Junjie and Hu, Weiming},
    booktitle = {The European Conference on Computer Vision (ECCV)}, month = {September}, year = {2018}
}

@InProceedings{track_kys,
    author={Bhat, Goutam and Danelljan, Martin and Van Gool, Luc and Timofte, Radu},
    title={Know Your Surroundings: Exploiting Scene Information for Object Tracking},
    booktitle = {The European Conference on Computer Vision (ECCV)}, month={August}, year={2020},
}

@inproceedings{raft,
    title={RAFT: Recurrent All-Pairs Field Transforms for Optical Flow},
    author={Teed, Zachary and Deng, Jia},
    booktitle = {The European Conference on Computer Vision (ECCV)}, month={August}, year={2020},
}

@inproceedings{lovasz_loss,
    author = {Berman, Maxim and Triki, Amal Rannen and Blaschko, Matthew B.},
    title = {The Lovász-Softmax Loss: A Tractable Surrogate for the Optimization of the Intersection-Over-Union Measure in Neural Networks},
    booktitle = {2018 IEEE/CVF Conference on Computer Vision and Pattern Recognition (CVPR)}, month={June}, year={2018}
}
